\documentclass[accepted]{melba}

\graphicspath{{figures/}}

\usepackage{microtype}
\usepackage{multirow}
\usepackage{booktabs}
\usepackage{array}
\newcolumntype{R}[1]{>{\hfill}p{#1}}
\usepackage[hidelinks]{hyperref}

\def\data#1{\vskip 0.3in\noindent{\color{melba}\large\bf Data Availability}\vskip 0.2in \noindent #1}

\melbaid{2025:004}  % This is provided by the publishing editor
\doi{10.59275/j.melba.2025-1a8b}
\melbaauthors{Danfeng Guo and Demetri Terzopoulos}  % Note: this one is also used to set the pdf 'authors' metadata
\email{lyleguo@g.ucla.edu}
\volume{3}
\firstpageno{59}  % Communicated by the publishing editor
\melbayear{2025}  % The publication year
\datesubmitted{2024-08-02}  % Date submitted to MELBA: mm/yyyy
\datepublished{2025-03-14}  % Today's date: mm/yyyy

\title{Prompting Medical Large Vision-Language Models to Diagnose\\ Pathologies by Visual Question Answering}

\ShortHeadings{Prompting MLVLMs to Diagnose Pathologies by VQA}{Guo and Terzopoulos}

\author{
	\name Danfeng Guo\aff{1},
	\name Demetri Terzopoulos\aff{1,2}}

\affiliations{% <- trailing '%' to avoid unwanted indent
	\num 1 \addr Computer Science Department, University of California, Los Angeles, CA, USA \\
	\num 2 \addr VoxelCloud, Inc., Los Angeles, CA, USA}

\abstract{%   <- trailing '%' for backward compatibility of .sty file
Large Vision-Language Models (LVLMs) have achieved significant success in recent years, and they have been extended to the medical domain. Although demonstrating satisfactory performance on medical Visual Question Answering (VQA) tasks, Medical LVLMs (MLVLMs) suffer from the hallucination problem, which makes them fail to diagnose complex pathologies. Moreover, they readily fail to learn minority pathologies due to imbalanced training data. We propose two prompting strategies for MLVLMs that reduce hallucination and improve VQA performance. In the first strategy, we provide a detailed explanation of the queried pathology. In the second strategy, we fine-tune a cheap, weak learner to achieve high performance on a specific metric, and textually provide its judgment to the MLVLM. Tested on the MIMIC-CXR-JPG and Chexpert datasets, our methods significantly improve the diagnostic F1 score, with the highest increase being 0.27. We also demonstrate that our prompting strategies can be extended to general LVLM domains. Based on POPE metrics, it effectively suppresses the false negative predictions of existing LVLMs and improves Recall by approximately 0.07.}


\begin{document}
\twocolumn[\maketitle]

\section{Introduction}

Research on Large Language Models (LLMs) has yielded astonishing results in recent years. LLMs with billions of parameters have achieved outstanding abilities in a wide range of application scenarios \citep{openai-chatgpt,openai2023gpt4,vicuna2023}. The success of LLMs has quickly extended into the Vision-Language (VL) domain. Large Vision-Language Models (LVLMs) are built upon LLMs by training adapters that project visual features into tokens that can be interpreted by LLMs \citep{li2023blip2,zhang2023llamaadapter,liu2023llava}. Visual Question Answering (VQA) is an essential skill of LVLMs, and VQA accuracy serves as a test metric for most of these models \citep{li2023blip2,zhang2023llamaadapter,zhu2023minigpt,liu2023llava}. LVLMs have been pretrained on medical datasets \citep{llavamed,medpalm,medpalm2} and they have been tested on medical VQA tasks \citep{vqarad,he2020pathvqa}. These Medical LVLMs (MLVLMs) have been able to answer questions regarding the imaging modalities, organs, and abnormalities depicted by the input medical scans.

However, ``hallucination'' has been a major problem for LVLMs. This refers to the generation of content that is contradictory to the input images. Hallucination can be measured via VQA. One may ask the model questions regarding the existence of objects in the input image(s) and the hallucination level is assessed as the percentage of correctly answered questions. VQA can also potentially serve for medical image diagnosis. Users pose questions regarding a pathology and the MLVLM responds based on its analysis of the medical scans. However, most of the available datasets involve simple questions such as ``what is the modality of this image'' and ``what is the organ/tissue in this image''. MLVLMs have yet to be thoroughly evaluated on VQA accuracy across a broad spectrum of pathologies. Additionally, general VQA models are usually tested by the commonly known accuracy: the percentage of correctly answered questions, which is an unsuitable measure for medical VQA.
Medical image classification metrics such as the Precision, Recall, and F1 are more suitable for the evaluation of medical VQA models. Several strategies have been explored to enhance the question answering of LLMs/LVLMs, including chain-of-thought prompting \citep{ddcot}, self-consistency \citep{wang-etal-2023-towards}, and retrieval-based augmentation \citep{caffagni2024wikillava}. All these methods involve fine-tuning the models, which is expensive. Training-free methods to improve the VQA accuracy of MLVLMs are desirable.

For MLVLMs, hallucination is exacerbated by imbalanced training data. Many pathologies are minority categories in medical datasets. Models trained on large-scale medical data may easily fail to learn the features of less common pathologies. Addressing data bias typically involves strategies such as including more data of better quality, but given the scarcity of medical data, significantly enlarging the dataset may not be feasible. Common remediations involve re-sampling the data such that the positive and negative cases are better balanced, but this poses challenges when the data involves multiple categories of pathology. Additionally, re-sampling may undermine the training needs of LVLMs, which generally require large quantities of data. These problems highlight the need of a cost-effective approach to navigate the problem of minority categories in datasets.

Our study focuses on the VQA abilities of MLVLMs. In particular, we test an existing MLVLM, LLaVA-Med \citep{llavamed} for chest X-ray VQA across 5 categories of pathologies. The results show that the model has low accuracy, especially on minority pathologies. To enhance its VQA accuracy, we propose two prompting strategies. The first involves enriching prompts with detailed explanations of the queried pathology. The explanations include how the queried pathology is defined and how it appears in images. Our second strategy involves introducing an auxiliary weak-learner model as another agent. We train a small image classifier and fine-tune it to identify negative images accurately. Then, the negative predictions of this classifier are appended to the prompt as a reference for the MLVLM.

We run our experiments on the MIMIC-CXR-JPG \citep{mimiccxr} and Chexpert \citep{chexpert} datasets. The results show that our prompt strategies improve the F1 score significantly in most pathology categories (highest +0.27). We also show that our weak-learner-prompting strategy is applicable to the general domain. It reduces the false negative predictions of general domain LVLMs and improves the Recall by around 0.07 according to POPE metrics \citep{li-etal-2023-evaluating}.

To summarize, our contributions include the following:
\begin{enumerate}
    \item We improve the VQA accuracy of MLVLMs by prompting with detailed explanations of pathologies.
    \item We introduce a low-cost weak learner model as a reference for LLaVA-Med, and this effectively reduces the false positive answers.
    \item We show that our second prompting strategy can be extended to general domains to help models adapt to specialized accuracy needs.
\end{enumerate}
\autoref{sec:related-work} reviews related work, \autoref{sec:methodology_cls} describes our methodology, \autoref{sec:exp_cls} presents our empirical study and its results, and \autoref{sec:concl-disc} draws conclusions from our research.

\section{Related Work}
\label{sec:related-work}

\paragraph{LVLMs and VQA}

LVLMs are built upon LLMs. A pretrained visual encoder extracts the visual features and an adapter module projects the extracted features to ones that can be understood by the LLM. Models of this type include those by \citet{liu2023llava}, \citet{zhu2023minigpt}, and \citet{zhang2023llamaadapter}. During training, the visual encoder and the LLM are usually fixed. VQA is an essential skill of LVLMs. Given an input image, the models should be able to answer questions correctly regarding that image.

\paragraph{Hallucination in LVLM VQA}

The hallucination problem usually refers to the LVLM generating a response that is not consistent with the input image. For VQA, in their generated answers the models may make mistakes on object presence, location, attributes, or the mutual relationship between objects. \citet{li-etal-2023-evaluating} find that frequently occurring objects are easily hallucinated by LVLMs, in that they tend to mention such objects even if it they are absent in the image. \citet{deceive} and \citet{deceive1} show that LVLMs sometimes presume the assumptions in questions are true and easily give wrong answers when asked about some objects not in the given image.

\paragraph{Causes of LVLM VQA Hallucination}

Hallucination can result from bias in the training data, missing fine-grained visual features, and LLM decoding strategies \citep{vllmsurvey}. For data bias, the imbalanced distribution of data is an important aspect. When most of the answers to a question in the training data are ``Yes'', the model tends to answer ``Yes'' to that question. Missing fine-grained visual features usually result from the pretraining of the visual encoder. Most LVLMs use the visual encoder of CLIP trained through contrastive learning. The encoder mainly focuses on salient features and ignores fine-grained features \citep{jain2023vcoder}. LVLM decoding strategies mostly choose the next word as the one having maximum conditional probability given previous text and the input image. This can lead to hallucination when the model overly relies on the knowledge learned in its training texts. Other causes include model simplicity and insufficient attention \citep{vllmsurvey}.

\paragraph{Mitigation of LVLM VQA Hallucination}

Strategies to mitigate hallucination in LVLMs mainly fall into two categories: prompt engineering and model improvement. Prompt engineering is a well developed technique in the natural language domain that provides LLMs with instructions and/or additional information to perform tasks. \citet{vu-etal-2024-freshllms} and \citet{peng2023checkfactstryagain} improve the LLM performance by providing external information. Regarding the former, \citet{deceive1} leverage visual instructions constructed from the bounding box information in the input image to prompt LLMs. \citet{ddcot} use a chain of thought scheme to prompt the models to perform step-by-step visual-language reasoning like humans, which eventually leads to the correct answers. \citet{wang-etal-2023-towards} generate multiple chains of thought and use the one with the majority vote as the answer. \citet{caffagni2024wikillava} prompt the model with explanations of the terms in questions. \citet{wang-etal-2024-soft-knowledge} retrieve external knowledge to assist the model in VQA tasks. With regard to the model improvement strategy for reducing hallucination, \citet{2023llavarlhf} improve the visual and text feature alignment through reinforcement learning. \citet{leng2023mitigating} propose a contrastive decoding strategy to reduce reliance on pretrained knowledge. \citet{favero2024multimodal} and \citet{zhao2024mitigating} also focus on the inference stage and propose specialized decoding strategies to mitigate hallucination. Other strategies for reducing hallucination have been proposed. For example, \citet{lure} design a post-processing model to detect hallucinated objects and rephrase the generated answers, and \citet{2023llavarlhf} adapt a reinforcement learning strategy that uses human evaluation of the hallucination level to improve the model.

\paragraph{Assessment of LVLM Hallucination}

There are two approaches to assessing hallucination in LVLMs. The first is VQA. The ground truth information of the input images is leveraged to construct questions regarding the existence of objects in the images (e.g., ``Is there a black cat in the image?''), as well as questions about objects which do not exist in the images. The models are evaluated in terms of the percentage of correctly answered questions. Metrics of this type include POPE \citep{li-etal-2023-evaluating}, CIEM \citep{hu2023ciem}, and NOPE \citep{lovenia2023negative}. The second  approach is to use pre-designed prompts from which the models produce various generations that are then evaluated. Examples include CHAIR \citep{rohrbach-etal-2018-object}, which counts the hallucinated objects in generated image captions, and MMHAL-BENCH \citep{2023llavarlhf}, which uses GPT-4 \citep{openai2023gpt4} to compare the generations with human answers and determine the propensity toward hallucination.

\paragraph{VQA in MLVLMs}

For MLVLMs, given a medical scan, models such as LLaVA-Med \citep{llavamed} and Med-PALM \citep{medpalm} are able to answer questions regarding the types of modalities, the scanned organs, and medical indicators such as opacity. They have demonstrated good performance on medical VQA datasets such as VQA-RAD \citep{vqarad}, SLAKE \citep{slake}, and Path-VQA \citep{he2020pathvqa}. XrayGPT \citep{Omkar2023XrayGPT} improved model performance by performing an additional round of training on selected high-quality data. CheXagent \citep{stanford-aimi-chexagent-2024} further developed the training process such that the visual encoder, projection layer, and the whole model are trained separately in three steps. However, most medical questions in existing datasets are simple (e.g., view classification). MLVLMs have not yet been tested on a broader range of complex pathologies. Recently, RaDialog \citep{pellegrini2023radialog} trains a separate image classifier and fine-tunes the MLVLM with the classification results on their designed datasets including various medical tasks. The image classifier helps the model generate medical reports with high medical correctness.

\begin{figure}
    \centering
    \includegraphics[width=\linewidth,trim={300 0 300 0},clip]{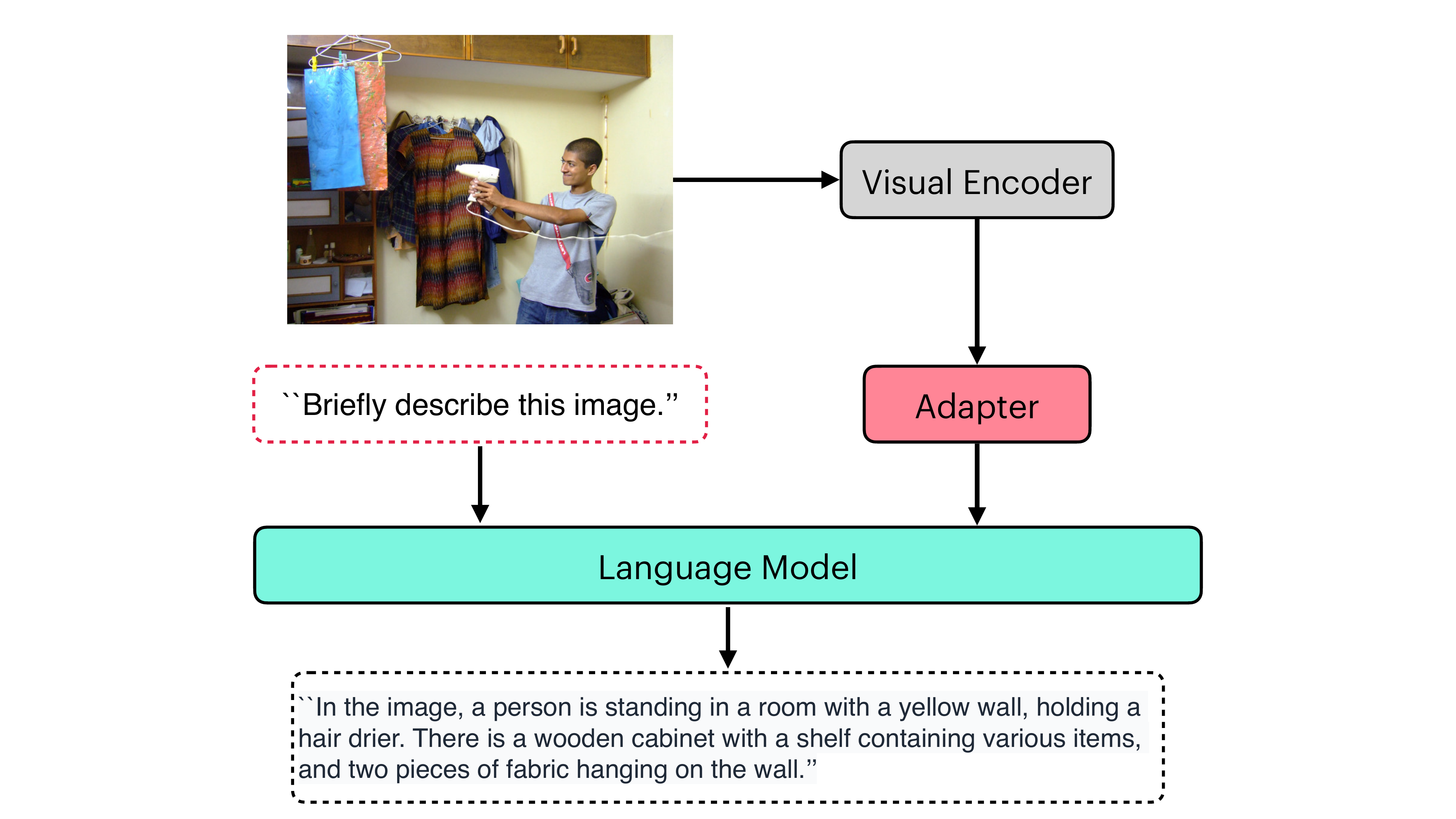}
    \caption{The structure of common LVLMs.}
    \label{fig:mllms}
\end{figure}

\paragraph{Medical Image Classification via VLMs}
The most notable vision-language model for medical image classification is ConVIRT \citep{convirt}, which employed a contrastive learning approach to pretrain the model for various tasks. It inspired CLIP \citep{clip}, which is able to utilize the pretrained model for zero-shot classification by searching for the best match between the image features and text features of disease categories. Several CLIP variants, such as BioMedCLIP \citep{biomedclip}, ChexZero \citep{tiu2022expert}, MedCLIP \citep{medclip}, Xplainer \citep{xplainer}, \cite{10.1007/978-3-031-16443-9_66}, and \cite{clipxray}, perform well on medical image classification tasks. They can also be fine-tuned for impressive performance on other tasks, such as segmentation and report generation. Compared with LVLMs, they are single-purpose models rather than generative AIs; however, these models may be used as the backbones of encoders in MLVLMs.

\section{Methodology}
\label{sec:methodology_cls}

\begin{figure}
    \centering
    \includegraphics[width=\linewidth,trim={100 60 0 0},clip]{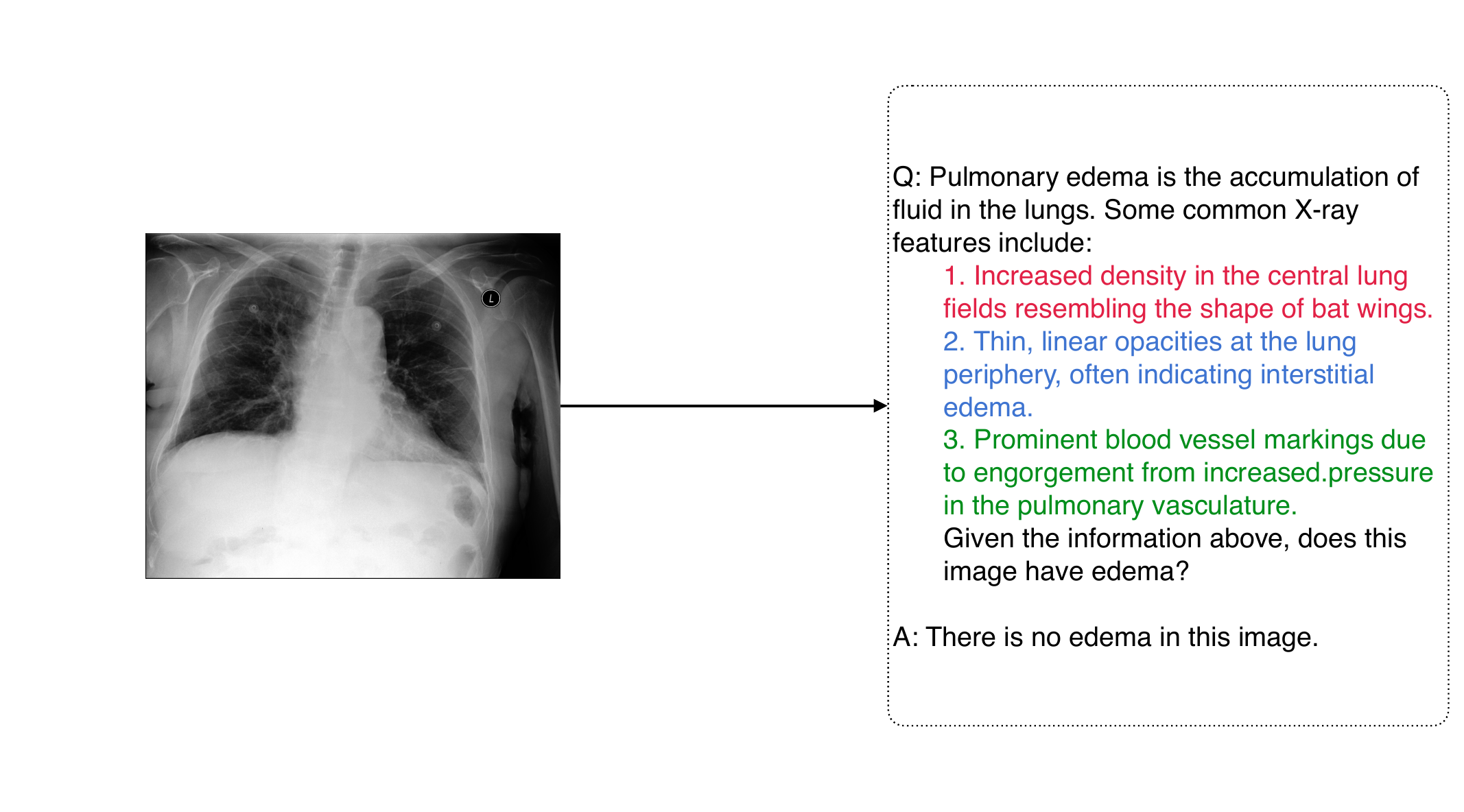}
    \caption{An example of including pathology explanations when prompting an MLVLM for medical VQA.}
    \label{fig:hintprompt}
\end{figure}

\autoref{fig:mllms} illustrates the structure of common LVLMs. They are based on a pretrained unimodal LLM such as Llama \citep{touvron2023llama} and Vicuna \citep{vicuna2023}. A pretrained visual encoder, such as ViT \citep{vit} or conventional CNNs, is applied to extract image features that are projected to the text feature space by an adapter. The projected visual features are concatenated with the text prompt embeddings and fed to the LLM. The adapter usually consists of several linear layers with non-linear activations. The visual encoder and the LLM are usually frozen during training.

In our work, we choose the pretrained LLaVA-Med \citep{llavamed} as our model, which is a MLVLM built upon LLaVA \citep{liu2023llava}. The model structure resembles \autoref{fig:mllms}. It uses pretrained Vicuna \citep{vicuna2023} as the LLM and the pretrained ViT encoder from CLIP \citep{clip} as the visual encoder. The adapter is simply a trainable projection matrix. Both the visual encoder and LLM weights are frozen during training. LLaVA-Med fine-tunes LLaVA in two steps. First, it fine-tunes LLaVA to generate medical reports from input medical images. Second, it uses GPT-4 to generate various questions from the ground truth reports and fine-tunes the model to perform question answering. 

Most MLVLMs are currently trained by medical VQA such that medical diagnosis can be performed by asking questions related to various pathologies; e.g., ``Does this image have lung lesion?''. To reduce model hallucination and improve VQA accuracy, we propose two prompting strategies at the inference stage: (1) providing the model with detailed explanations about the queried pathologies and (2) asking the model to consider the inferences of a weak learner.

\subsection{Prompting With Detailed Explanations}

Given imbalanced training data, MLVLMs might not adequately be able to learn the features of the minority pathologies. To compensate for insufficient training, we provide a detailed explanation of the queried pathology as a prompt at the inference stage. The explanation briefly defines the pathology and lists several key findings in medical images that may indicate its existence. An example is shown in \autoref{fig:hintprompt}. The model is informed that Pulmonary Edema is defined as the accumulation of fluid in the lungs. Then several chest X-ray findings that may suggest its existence are provided. The model can determine if the given image has Pulmonary Edema by linking the given findings with the image features.

Prompt templates for a number of pathologies are listed in \autoref{app:append_prompts}. 

\begin{figure}
    \centering
    \includegraphics[width=\linewidth,trim={100 60 0 0},clip]{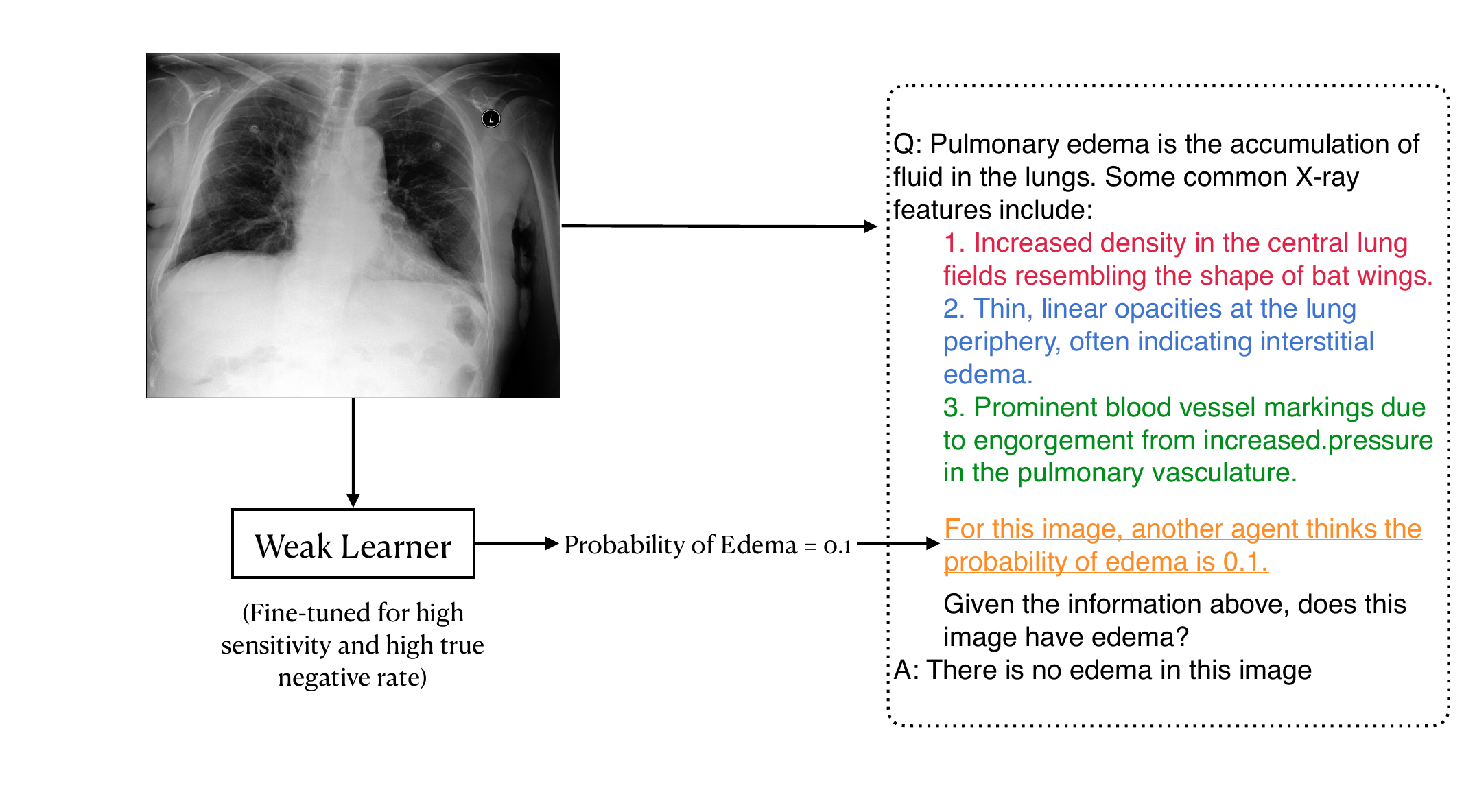}
    \caption{An example of prompting an MLVLM for medical VQA using both pathology explanations and reference predictions from a weak learner.}
    \label{fig:hintprompt2}
\end{figure}

\subsection{Prompting With Detailed Explanations and Weak Learners}

Data re-sampling is a commonly-used strategy to deal with imbalanced datasets that are responsible for the tendency of traditional image classification models to return negative predictions for minority pathologies. Models trained on re-sampled datasets often exhibit improvements in Precision and Recall scores; however, this strategy may not be suitable to MLVLMs for two reasons. First, it is difficult to balance a dataset containing many categories of pathologies. Second, MLVLMs usually demand much larger datasets and fine-tuning is also expensive.

One can nevertheless enable MLVLMs to benefit by leveraging small models trained on re-sampled datasets. Our method resembles multiagent LLM systems, such as \citet{du2023improving}, where multiple LLMs debate each other and hallucination can be corrected by referring to the generated outputs of other models. Given that traditional image classifiers are smaller, it is feasible to train multiple small classifiers each of which is trained on re-sampled datasets of a particular pathology. Those models can be further fine-tuned to optimize a single aspect, such as fewer False Positives (FPs) or fewer False Negatives (FNs). The classifiers are applied to the medical images and return preliminary predictions. These predictions are selectively included in the prompts as references for the MLVLM. Hence, MLVLMs can benefit indirectly from the nuanced understanding that these specialized models can provide. This method is meaningful because clinicians usually must balance the trade-off between overtreatment and undertreatment when making healthcare decisions. For instance, they may prefer models having a low FP rate if the cost of overtreatment is higher than that of undertreatment. 

 An example is shown in \autoref{fig:hintprompt2}, which queries about the presence of Edema. We first provide the model with the detailed explanation of Edema. Then, we use the weak learner to suppress the FPs. The image is input to an Edema classifier that has been fine-tuned on a balanced dataset for high sensitivity and high true negative (TN) rate. If its prediction is negative, we append after the pathology explanation the prompt ``For this image, another agent thinks the probability of Edema is 0.1''. Instead of using the actual predicted probability, the probability value is manually chosen because the decision threshold has been fine-tuned and is no longer 0.5. We do not use a zero probability value because we do not want the model overly to trust the weak learner. Although in this example our goal is only to reduce FPs, our strategy can also be applied to reduce FNs, simply by fine-tuning the classifier for a high True Positive (TP) rate and applying the prompt in the case of positive predictions.

 \section{Empirical Study}
\label{sec:exp_cls}

\subsection{Datasets}
\label{sec:datasets}

\begin{table}
    \centering
    \begin{tabular}{l r}
    \toprule
         Pathology & $+$Cases \\
         \midrule
        Atelectasis & 64\\
        Cardiomegaly & 31\\
        Consolidation & 335\\
        Edema & 1,276\\
        Pleural Effusion & 260\\
    \bottomrule
    \end{tabular}
    \caption{Positive ($+$) case counts for the 5 test pathologies in the LLaVA-Med training set.}
    \label{tab:pmc15}
\end{table}

LLaVA-Med is pretrained on the PMC-15M dataset \citep{pmc15}, which contains image-text pairs of multiple modalities; e.g., CT, MRI, X-ray, etc. In the first stage, 467,710 image-report pairs were selected for training. In the second stage, 56,708 question-answer pairs were created from the data of the first stage to fine-tune the model. \autoref{tab:pmc15} shows the count of reports in the LLaVA-Med training data (second stage) that mention one of the five test pathologies as positive. Relative to the total amount of data, all five categories are minorities.

\begin{table*}
    \centering
    \setlength{\tabcolsep}{4pt}
    \begin{tabular}{l *{4}{R{20mm}}}
    \toprule
    & \multicolumn{2}{c}{MIMIC-CXR-JPG (5,159)} & \multicolumn{2}{c}{Chexpert (668)} \\
    \cmidrule(lr){2-3} \cmidrule(lr){4-5}
        Category & $+$Cases & $-$Cases & $+$Cases & $-$Cases \\
    \midrule
        Atelectasis & 1,034 & 4,125 & 178 & 490\\
        Cardiomegaly & 1,258 & 3,901 & 175 & 493\\
        Consolidation & 326 & 4,833 & 35 & 633\\
        Edema & 959 & 4,200 & 85 & 583\\
        Enlarged Cardiomediastinum & 200 & 4,959 & 298 &370\\
        Fracture & 167 & 4,992 & 6 & 662\\
        Lung Lesion & 202 & 4,957 & 14 & 654\\
        Lung Opacity & 1,561 & 3,598 & 310 & 358\\
        Pleural Effusion & 1,542 & 3,617 & 120 & 548\\
        Pleural Other & 119 & 5,040 & 8 & 660\\
        Pneumonia & 539 & 4,620 & 14 & 654\\
        Pneumothorax & 144 & 5,015 & 10 & 658\\
        Support Devices & 1,457 & 3,702 & 315 & 353\\
        \bottomrule
    \end{tabular}
    \caption{Splits of positive ($+$) and negative ($-$) cases (`uncertain' is regarded as negative) for the 13 finding categories in the MIMIC-CXR-JPG and Chexpert test sets.}
    \label{tab:datasplit}
\end{table*}

To assess the zero-shot performance of the MLVLM, we used the MIMIC-CXR-JPG \citep{mimiccxr} and Chexpert \citep{chexpert} chest X-ray test sets. They include 5,159 and 668 images, respectively. 
Neither dataset overlaps with PMC-15M.

MIMIC-CXR-JPG includes images and medical reports covering 13 categories of findings: Atelectasis, Cardiomegaly, Consolidation, Edema, 
Enlarged Cardiomediastinum, Fracture, Lung Lesion, Lung Opacity, Pleural Effusion, Pneumonia, Pneumothorax, Pleural Other, and Support Devices. The raw reports are parsed and rough image-level tags are automatically generated by a rule-based approach \citep{chexpert}. Each label contains four values: 1 (positive), 0 (negative), $-1$ (uncertain), and missing. For simplicity, we treat both uncertain and missing as negative. We also use the MIMIC-CXR-JPG training set, which contains 227,827 chest X-rays with reports, to train the weak learner models.

Chexpert covers the same 13 categories as MIMIC-CXR-JPG. However, it does not include medical reports and has only image-level labels. There is no overlap between MIMIC-CXR-JPG and Chexpert. 

\autoref{tab:datasplit} shows the split of pathology categories (excluding normal) in the MIMIC-CXR-JPG and Chexpert test sets. Clearly, almost all pathology categories are minor classes with much fewer positive than negative occurrences.

For our main testing regimen, we selected the five pathologies in the Chexpert Competition \citep{chexpert}: Atelectasis, Cardiomegaly, Consolidation, Edema, and Pleural Effusion.

\subsection{Implementation Details}
\label{sec:implementation}

\begin{table*}[t!]
    \centering
    \begin{tabular}{r p{13cm}}
    \toprule
        Prompt & Template \\
        \midrule
        PT1 &  ``Does this image have \{target\}?'' \\[10pt]
        PT2 & ``\{explanation\} Given the information above, does this image have \{target\}?'' \\[10pt]
        PT3 & ``\{explanation\} For this image, another agent thinks the probability that it has \{target\} is \{n\} percent. Given the information above, does this image have \{target\}?'' \\
    \bottomrule
    \end{tabular}
    \caption{The Prompt Templates (PTs). \{target\} is the pathology cited in the questions. \{explanation\} contains a pathology explanation among those listed in \autoref{app:append_prompts}. \{n\} is the probability associated with the weak learner.}
    \label{tab:prompt_template}
\end{table*}

As was mentioned in \autoref{sec:methodology_cls}, we use the pretrained LLaVA-Med MLVLM without any further fine-tuning. We convert the classification task into a VQA task by using the prompt template shown in Row 1 of \autoref{tab:prompt_template}, which we name Prompt Template~1 (PT1). We first run the pretrained LLaVA-Med with PT1. Next, we incorporate pathology explanations (Row 2 of \autoref{tab:prompt_template}), yielding Prompt Template~2 (PT2). Finally, we integrate the predictions of weak learners into the prompts (Row 3 of \autoref{tab:prompt_template}), resulting in Prompt Template~3 (PT3).

As will be justified by our experiments, our weak learner is designed to suppress FP predictions. To this end, we use the pretrained ResNet50 \citep{resnet}. Given the low cost of the weak learner, we train a model separately for each pathology with each training dataset sampled such that the ratio of positive and negative cases is $2:1$. The model was trained for 10 epochs with a $1e-4$ learning rate. The training process was monitored using the AUC score and the one with the highest validation AUC was kept. Then, the decision threshold $d$ was fine-tuned to optimize a weighted sum of Specificity and Negative Predictive Value (NPV); i.e.,
\begin{equation}
    d = w_1\frac{\text{TN}}{\text{TN}+\text{FP}}+w_2\frac{\text{TN}}{\text{TN}+\text{FN}},
\end{equation}
where weights $w_1$ and $w_2$ are preset to 0.2 and 0.8, respectively. The medical images were input to the weak learners to obtain preliminary predictions for each pathology and only the negative predictions were selected to craft the PT3 prompts. 

The responses returned by LLaVA-Med can take various forms, such as ``This image has Edema'', ``Edema is found'', ``The fluid in the lung indicates Edema'', etc. An off-the-shelf Llama-7B \citep{touvron2023llama} serves to summarize long responses into Yes/No answers such that accuracies could easily be computed.

\subsection{Results}

To demonstrate the efficacy of our prompting strategies, starting from the PT1 baseline, the pathology explanations were provided first (strategy PT2) and then, based on the results, weak learners were introduced to improve performance on specific aspects, resulting in strategy PT3.

\begin{table*}
    \centering
    \begin{tabular}{l l *{4}{R{13mm}}}
    \toprule
    & & \multicolumn{2}{c}{MIMIC-CXR-JPG} & \multicolumn{2}{c}{Chexpert}\\
    \cmidrule(lr){3-4} \cmidrule(lr){5-6}
       Pathology & Metric & PT1 & PT2 & PT1 & PT2 \\
         \midrule
        \multirow{3}{*}{Atelectasis} & Precision & 19.5 & 20.0 & 30.5 & 26.5\\
        & Recall & 41.5 & 92.9 & 44.4 & 91.6 \\
        & F1 & 26.5 & 33.0 & 36.5 & 41.0 \\
        \midrule
        \multirow{3}{*}{Cardiomegaly} & Precision & 25.8 & 24.6 & 27.1 & 26.0\\
        & Recall & 22.5 & 89.4 & 20.0 & 86.3 \\
        & F1 & 24.0 & 38.6 & 23.0 & 40.0 \\
        \midrule
        \multirow{3}{*}{Consolidation} & Precision & 6.8 & 6.3 & 6.0 & 5.2 \\
        & Recall & 42.3 & 98.5 & 40.0 & 97.1 \\
        & F1 & 11.7 & 11.9 & 10.4 & 9.8 \\
        \midrule
        \multirow{3}{*}{Edema}& Precision & 19.6 & 18.5 & 11.7 & 13.7 \\
        & Recall & 36.0 & 72.7 & 29.4 & 76.5 \\
        & F1 & 25.4 & 29.5 & 16.8 & 23.2 \\
        \midrule
        \multirow{3}{2cm}{Pleural Effusion}& Precision & 30.4 & 30.0 & 22.3 & 17.9 \\
        & Recall & 42.8 & 92.7 & 49.2 & 90.0 \\
        & F1 & 35.6 & 45.3 & 30.7 & 29.9 \\
        \bottomrule
    \end{tabular}
    \caption{LLaVA-Med VQA performance evaluated by Precision, Recall, and F1 (\%) scores of five pathologies on the MIMIC-CXR-JPG and Chexpert test datasets.}
    \label{tab:llmqa_mi}
\end{table*}

\paragraph{PT2: Adding Pathology Explanations}
\autoref{tab:llmqa_mi} reports Precision, Recall, and F1 scores of the PT1 and PT2 strategies on the MIMIC-CXR-JPG and Chexpert test sets.\footnote{The AUC and ROC scores commonly reported in the literature to assess the performances of most medical image classification models on the MIMIC-CXR-JPG and Chexpert datasets are unsuitable in our context because MLVLMs output text rather than probabilities.} On MIMIC-CXR-JPG, after adding pathology explanations, the F1 scores increased for detecting Atelectasis, Cardiomegaly, Edema, and Pleural Effusion, albeit only minimally for Consolidation. On Chexpert, after adding pathology explanations, the F1 scores for detecting Atelectasis, Cardiomegaly, and Edema increased, whereas they did not for Consolidation and Pleural Effusion. The Precision and Recall scores reveal that adding explanations generally leads to a large increase in Recall, but only minimally influences Precision. For minority pathologies such as Consolidation whose F1 score is dominated by low Precision, improving the Recall would not have much effect. Thus, PT2's performance bottleneck is Precision.

\begin{table}
    \centering
    \begin{tabular}{l c c c}
    \toprule
         Pathology& TP & FP & FN\\
         \toprule
        Atelectasis & 163 & 453 & 15 \\
        Cardiomegaly & 151 & 430 & 24 \\
        Consolidation & 28 & 557 & 7 \\
        Edema & 65 & 410 & 20 \\
        Pleural Effusion & 108 &495 & 12 \\
        \bottomrule
    \end{tabular}
    \caption{True positive (TP), false positive (FP), and false negative (FN) counts of LLaVA-Med with the PT2 strategy on the Chexpert test set.}
    \label{tab:TPFP_count}
\end{table}

\begin{table}
    \centering
    \begin{tabular}{l c c c c}
      \toprule 
       Pathology & AUC & Precision & {Recall} & {F1}\\
         \midrule
        Atelectasis & 82.2 & 62.0 & {56.7} & {59.2}\\
        Cardiomegaly & 85.0 & 74.4 & {38.3} & {50.6}\\
        Consolidation & 81.7 & 100 & {2.9} & {5.6}\\
        Edema & 87.3 & 46.5 & {61.2} & {54.2}\\
        Pleural Effusion & 91.2 & 36.6 & {94.2} & {52.7}\\
        \bottomrule
    \end{tabular}
    \caption{Performance of the weak learner on the Chexpert test sets for the 5 pathologies.}
    \label{tab:resnet_result}
\end{table}

\paragraph{PT3: Referring to Weak Learners}
Going beyond our PT2 strategy, we applied our PT3 strategy to further improve diagnostic accuracy. \autoref{tab:TPFP_count} provides the TP, FP, and FN prediction counts of LLaVA-Med on the Chexpert test set using the PT2 strategy. Note the large number of FP cases. Hence, we designed our weak learners to suppress FP predictions. As mentioned in \autoref{sec:implementation}, we trained the model for the highest AUC and then fine-tuned the decision threshold. \autoref{tab:resnet_result} reports the AUC, Precision, Recall and F1 (after fine-tuning) of the weak learner. It is important to note that we fine-tuned the decision threshold to achieve high specificity and negative predictive value; thus, the reported Precision, Recall, and F1 scores are based on this fine-tuned threshold and may not be directly comparable to those of other models. 
\autoref{tab:ref_prompt} compares the performance on Chexpert before and after referring to the weak learner. It shows that the F1 prediction accuracy can be substantially increased by introducing weak learner predictions into the prompts. The F1 scores of Cardiomegaly, Edema, and Pleural Effusion increase by 0.115, 0.194 and 0.089, respectively. To further demonstrate the efficacy of our PT3 strategy, \autoref{tab:TPFP_change} compares the FP predictions of the PT2 and PT3 strategies. The reduction of FP cases is noteworthy, especially on Edema, for which the FP count is reduced by 78.5\% (322). 

\begin{table}
    \centering
    \begin{tabular}{l l c c }
    \toprule
       Pathology & Metric & PT2 & PT3\\
         \midrule
        \multirow{3}{*}{Atelectasis}& Precision  & 26.5 & 28.8 \\
        & Recall  & 91.6 & 83.1 \\
        & F1  & 41.0 &  42.8 \\
        \midrule
        \multirow{3}{*}{Cardiomegaly}& Precision  & 26.0 &  38.1\\
        & Recall & 86.3 &  79.4\\
        & F1 & 40.0 & 51.5\\
        \midrule
        \multirow{3}{*}{Consolidation}& Precision & 5.2 &  7.5\\
        & Recall &  97.1 &  34.3\\
        & F1 & 9.8 & 12.2\\
        \midrule
        \multirow{3}{*}{Edema}& Precision  & 13.7 &  36.8 \\
        & Recall & 76.5 &  50.6\\
        & F1 & 23.2 & 42.6\\
        \midrule
        \multirow{3}{2cm}{Pleural Effusion}& Precision &  17.9 &  25.0\\
        & Recall &90.0 &  85.0\\
        & F1 & 29.9 & 38.8\\
        \bottomrule
    \end{tabular}
    \caption{Diagnostic accuracies of LLaVA-Med with the PT2 and PT3 strategies on the Chexpert test set.}
    \label{tab:ref_prompt}
\end{table}

\paragraph{Additional VQA Experiments}
\autoref{tab:llmqa_mi_chx_add} shows the results of applying the PT1, PT2, and PT3 strategies with LLaVA-Med on the MIMIC-CXR-JPG and Chexpert datasets across another five medical findings: Enlarged Cardiomediastinum, Lung Lesion, Lung Opacity, Pneumonia, and Pneumothorax. Providing pathology explanations (PT2) generally yields better results over the PT1 baseline, albeit inconsistently. Introducing weak learner references (PT3) yields only limited increases in Precision, but large decreases in Recall. Generally, it offers insignificant improvement. Enlarged Cardiomediastinum, Lung Lesion, Pneumonia, and Pneumothorax are minor categories and all our experimental settings, including for the weak learner, fail to learn them. Prompting is apparently unhelpful in such situations.

\paragraph{SOTA Benchmark}
\citet{tiu2022expert} report F1 scores for detecting Atelectasis, Cardiomegaly, Consolidation, Edema, and Pleural Effusion on the Chexpert dataset using their deep learning model, as well as for the performance of radiologists. Their work offers a state-of-the-art chest X-ray diagnosis benchmark. \autoref{tab:sota_comp_cls1} compares the F1 scores of radiologists, the model of \citet{tiu2022expert}, and LLaVA-Med. It shows that LLaVA-Med's VQA performance of with the baseline PT1 strategy is unsatisfactory, rendering the model far from being deployable in clinical practice. However, while still underperforming radiologists, our PT3 strategy yields a significant improvement, especially on Atelectasis, Cardiomegaly, and Edema for which the F1 score increases by approximately 17\% to 21\%. 

\begin{table}
    \centering
    \begin{tabular}{l c c}
    \toprule
    Pathology & PT2 & PT3\\
         \midrule
        Atelectasis & 453 &  365\\
        Cardiomegaly & 430 & 226\\
        Consolidation & 557 & 149 \\
        Edema & 410 & 88 \\
        Pleural Effusion & 495 & 304 \\
        \bottomrule
    \end{tabular}
    \caption{False positive counts of LLaVA-Med with the PT2 and PT3 strategies on the Chexpert test set.}
    \label{tab:TPFP_change}
\end{table}

\begin{table*}
    \centering
    \begin{tabular}{l l *{6}r}
      \toprule
      & & \multicolumn{3}{c}{MIMIC-CXR-JPG} & \multicolumn{3}{c}{Chexpert} \\
      \cmidrule(lr){3-5} \cmidrule(lr){6-8}
       Pathology & Metric & PT1 & PT2 & PT3 & PT1 & PT2 & PT3 \\
         \midrule
        \multirow{3}{*}{Enlarged Cardiomediastinum} & Precision & 4.3 & 3.9 & 5.0 & 49.3 & 44.1 & 49.4\\
        & Recall & 15.0 & 89.0 & 53.5 & 12.4 & 85.2 & 59.1 \\
        & F1 & 6.7 & 7.4 & 9.1 & 19.8 & 58.1 & 53.8 \\
        \midrule
        \multirow{3}{*}{Lung Lesion} & Precision & 3.9 & 3.9 & 4.2 & 2.0 & 2.1 & 2.9\\
        & Recall & 77.2 & 100.0 & 66.3 & 71.4 & 100.0 & 92.9\\
        & F1 & 7.4& 7.5 & 8.0 & 3.9& 4.1 & 5.7 \\
        \midrule
        \multirow{3}{*}{Lung Opacity} & Precision & 31.4 & 30.4 & 31.9 & 50.0 & 47.2 & 51.4\\
        & Recall & 67.6& 88.8 & 84.1 & 70.3 & 90.7 & 84.8 \\
        & F1 & 42.8& 45.3 & 46.2 & 58.5 & 62.1 & 63.3\\
        \midrule
        \multirow{3}{2cm}{Pneumonia} & Precision & 11.4 & 10.5 & 12.3 & 2.4 & 1.7 & 6.3 \\
        & Recall & 20.0 & 74.6 & 18.0 & 21.4 & 57.1 & 28.6\\
        & F1 & 14.6 & 18.4 & 14.6 & 4.4 & 3.4 & 10.4\\
        \midrule
        \multirow{3}{2cm}{Pneumothorax} & Precision & 3.0 & 2.6 & 3.6 & 0.0 & 1.7 & 2.0\\
        & Recall & 16.7 & 78.5 & 50.7 & 0.0 & 90.0 & 50.0\\
        & F1 & 5.1 & 5.1 & 6.8 & 0.0 & 3.3 & 3.8\\
        \bottomrule
    \end{tabular}
    \caption{LLaVA-Med VQA performance on the MIMIC-CXR-JPG and Chexpert test sets for another 5 pathologies.}
    \label{tab:llmqa_mi_chx_add}
\end{table*}

\begin{table*}[t!]
    \centering
    \begin{tabular}{l c c c c}
    \toprule
         Pathology &  Radiologist & \citep{tiu2022expert} & PT1 & PT3 \\
         \midrule
        Atelectasis & 69.2 & 64.6 & 26.5 & 41.3\\
        Cardiomegaly & 67.8 & 74.3 & 24.0 & 51.5\\
        Consolidation & 38.5 & 33.3 & 11.7 & 12.2\\
        Edema & 58.3 & 60.2 & 25.4 & 42.6\\
        Pleural Effusion & 73.7 & 70.4 & 35.5 & 46.8\\
        \bottomrule
    \end{tabular}
    \caption{F1 scores (\%) on 5 pathologies in the Chexpert test set, including for the radiologist diagnoses, the state-of-the-art benchmark \citep{tiu2022expert}, as well as LLaVA-Med VQA with the PT1 scenario and the PT3 scenario, which is the best result achieved by applying both our prompting strategies.}
    \label{tab:sota_comp_cls1}
\end{table*}

\begin{table*}[t!]
  \centering
    \begin{tabular}{l *{9}c}
    \toprule
         & \multicolumn{3}{c}{POPE Adversarial} & \multicolumn{3}{c}{POPE Popular} & \multicolumn{3}{c}{POPE Random} \\
         \cmidrule(lr){2-4} \cmidrule(lr){5-7} \cmidrule(lr){8-10}
         Model & Precision & Recall & F1 & Precision & Recall & F1 & Precision & Recall & F1 \\
         \midrule
         LLaVA & 91.0 & 78.8 & 84.5 & 95.2 & 78.8 & 86.2 & 97.4 & 78.8 & 87.1 \\
         with referral & 88.4 & 85.7 & 87.0 & 92.8 & 85.7 & 89.0 & 97.3 & 85.7 & 91.1 \\
         \midrule
         MiniGPT-v2 & 88.2 & 77.2 & 82.3 & 92.7 & 77.2 & 84.2 & 97.2 & 77.2 & 86.1 \\
         with referral & 86.8 & 84.2 & 85.5 & 91.9 & 84.2 & 87.9 & 97.3 & 84.2 & 90.3 \\
         \bottomrule
    \end{tabular}
    \caption{Comparison of POPE scores for LVLM models with and without referring to the predictions of weak learners.}
    \label{tab:general_mllm}
\end{table*}

\paragraph{Application to General Domain LVLMs}
Our prompt strategies can also be applied to general domain LVLMs. We studied the performance of LLaVA \citep{liu2023llava} and MiniGPT-v2 \citep{zhu2023minigpt} using POPE metrics \citep{li-etal-2023-evaluating}. POPE evaluates the hallucination of LVLMs by asking questions about the existing/non-existing objects on given images. The input images are from MSCOCO dataset and there are three questions categories: Random (ramdom sample objects for questions), Popular (frequent objects), and Adversarial (frequent but non-existent objects). The performance is evaluated by the Precision, Recall and F1 of correctly answered questions. The POPE scores of LLaVA and MiniGPT-v2 have high Precision and low Recall. Hence, our weak learner strategy can be used to reduce the FN predictions. We selected an off-the-shelf Fast-RCNN \citep{girshick2015fast} as the weak learner, fine-tuned the detection threshold of bounding box scores to achieve high Recall, and introduced the positive predictions of the weak learner into the prompts. The results in \autoref{tab:general_mllm} show that the Recall scores across three POPE categories increased by around 7\% (Precision scores decrease slightly), thus improving the F1 scores.

\section{Conclusions and Discussion}
\label{sec:concl-disc}

We have tested the visual question answering abilities of the LLaVA-Med medical large vision-language model when applied to the diagnosis of pathologies. Our results show that the model has unsatisfactory performance when asked questions regarding the presence of complex pathologies. We proposed two prompt engineering strategies to improve the visual question answering accuracy of the model: providing explanations of pathologies and referring to the predictions of weak learners. The first strategy helps the model understand minority pathologies that it does not learn well in the training stage. The second strategy can help improve diagnostic accuracy in specific ways; e.g., by suppressing false positives. This strategy can also be applied to LVLMs in other, non-medical domains.

However, our two strategies are not effective on pathologies with extremely scarce data. For example, providing text explanations for Consolidation, Fracture, Lung Lesion, Pneumonia, and Pneumothorax may not suffice since the visual encoder does not adequately learn meaningful visual features. Moreover, the data may not suffice to adequately train weak learners. A promising direction for future re-
{\parfillskip=0pt\par}\newpage\noindent % balance this page
search would be to devise a strategy for handling these rare categories. Retrieval Augmented Generation (RAG) could be a potential solution. For instance, in addition to textual explanations of pathologies, typical example images can be provided to help the model make diagnostic decisions.

\clearpage

\acks{DG was supported in part by an unrestricted donation to UCLA from VoxelCloud, Inc.}

\ethics{The work follows appropriate ethical standards in conducting research and writing the manuscript, following all applicable laws and regulations regarding treatment of animals or human subjects.}

\coi{We have no conflicts of interest.}

\data{Only publicly available datasets were used.}

\appendix

\section{Pathology Explanations}
\label{app:append_prompts}

The explanations of the five pathologies are as follows:

\paragraph{Atelectasis:} Atelectasis refers to the partial or complete collapse of a lung or a section of lung. The features of atelectasis on an X-ray can vary depending on the cause and extent of the collapse. Some common X-ray features include: 1.~The affected area may appear denser or whiter than normal lung tissue due to the collapse, leading to increased opacity on the X-ray. 2.~The affected portion of the lung may appear smaller or compressed compared to the surrounding healthy lung tissue. 3.~Atelectasis can cause a shift or displacement of nearby structures, such as the trachea or heart, toward the affected area. 4.~In obstructive atelectasis (caused by a blockage in the airways), there might be signs of hyperinflation in the unaffected areas of the lung and a visible blockage or narrowing in the affected bronchus. 5.~Linear or band-like opacities may be visible, often referred to as plate or band atelectasis, which can occur due to the collapse of small airways. Given the information above, does this image have Atelectasis?

\paragraph{Cardiomegaly:} Cardiomegaly is enlargement of the heart. The definition is when the transverse diameter of the cardiac silhouette is greater than or equal to 50\% of the transverse diameter of the chest (increased cardiothoracic ratio) on a posterior-anterior projection of a chest radiograph or a computed tomography. Given the information above, does this image have Cardiomegaly?

\paragraph{Consolidation:} Consolidation on an X-ray refers to the filling of the lung's air spaces with fluid inflammatory exudate, or cellular material. Typical X-ray findings suggesting consolidation include: 1.~Areas of increased density in the lung tissue, appearing as an opaque or hazy patch on the X-ray. Given the information above, does this image have Consolidation?

\paragraph{Edema:} Pulmonary edema is the accumulation of fluid in the lungs. Some common X-ray features include: 1.~Increased density in the central lung fields resembling the shape of bat wings. 2.~Thin, linear opacities at the lung periphery, often indicating interstitial edema. 3.~Prominent blood vessel markings due to engorgement from increased pressure in the pulmonary vasculature. Given the information above, does this image have Edema?

\paragraph{Pleural Effusion:} Pleural effusion is the accumulation of fluid in between the parietal and visceral pleura. Some common X-ray features include: 1.~blunting of the costophrenic~/ cardiophrenic angle. 2.~fluid within the horizontal or oblique fissures. 3.~meniscus is seen. 4.~mediastinal shift occurs away from the effusion.

\end{document}